\documentclass[journal]{IEEEtran}

\ifCLASSINFOpdf
\else
   \usepackage[dvips]{graphicx}
\fi
\usepackage{url}

\usepackage{graphicx}
\usepackage{bbding}
\usepackage[table,xcdraw]{xcolor}
\usepackage{amsmath,amssymb,amsfonts}
\usepackage[numbers,sort&compress]{natbib}
\usepackage{threeparttable}
\usepackage{amsthm}
\usepackage{arydshln}
\usepackage{mwe}
\usepackage{lipsum}

\usepackage{multirow}
\usepackage{adjustbox}
\usepackage{tikz}

\begin{document}

\title{Image Super-Resolution with Taylor Expansion Approximation and Large Field Reception}

\author{Jiancong Feng, Yuan-Gen Wang, Mingjie Li, and Fengchuang Xing

\IEEEcompsocitemizethanks{

\IEEEcompsocthanksitem Feng, J., Wang, Y.-G., Li, M., and Xing F. are with the School of Computer Science and Cyber Engineering, Guangzhou University, Guangzhou 510006, China (e-mail: fengjiancong@e.gzhu.edu.cn; wangyg@gzhu.edu.cn; limingjie@gzhu.edu.cn; xfchuang@e.gzhu.edu.cn). \protect\\
}

}


\markboth{}
{Shell \MakeLowercase{\textit{et al.}}: Bare Demo of IEEEtran.cls for IEEE Journals}
\maketitle

\begin{abstract}
Self-similarity techniques are booming in blind super-resolution (SR) due to accurate estimation of the degradation types involved in low-resolution images. However, high-dimensional matrix multiplication within self-similarity computation prohibitively consumes massive computational costs. We find that the high-dimensional attention map is derived from the matrix multiplication between Query and Key, followed by a softmax function. This softmax makes the matrix multiplication between Query and Key inseparable, posing a great challenge in simplifying computational complexity. To address this issue, we first propose a second-order Taylor expansion approximation (STEA)  to separate the matrix multiplication of Query and Key, resulting in the complexity reduction from $\mathcal{O}(N^2)$ to $\mathcal{O}(N)$. Then, we design a multi-scale large field reception (MLFR) to compensate for the performance degradation caused by STEA. Finally, we apply these two core designs to laboratory and real-world scenarios by constructing LabNet and RealNet, respectively. Extensive experimental results tested on five synthetic datasets demonstrate that our LabNet sets a new benchmark in qualitative and quantitative evaluations. Tested on the RealWorld38 dataset, our RealNet achieves superior visual quality over existing methods. Ablation studies further verify the contributions of STEA and MLFR towards both LabNet and RealNet frameworks.
\end{abstract}


\IEEEpeerreviewmaketitle

\section{Introduction}

\IEEEPARstart{S}{ingle}-Image Super-Resolution (SISR) \cite{1,2,3,4} has obtained widespread application in various fields, such as remote sensing imagery, security surveillance, and medical imaging. SISR methods are generally divided into two categories: non-blind SR and blind SR. The non-blind SR \cite{tmm_large,tmm_dynamic,tmm_accurate,tmm_iterative,tmm_cross,ESPCN,rcan} assumes that the degradation type is known, which degradation function is typically represented as:
\begin{equation}
\label{eq:non-blind}
\mathbf{x}_{lr} = \mathbf{x}_{hr} \downarrow_{s},
\end{equation}
where $\mathbf{x}_{hr}$ and $\mathbf{x}_{lr}$ denote the HR and LR images respectively, and $\downarrow_{s}$ denotes the bicubic downsampling operation with a scale factor of $s$. However, the non-blind SR represented by Eq. (\ref{eq:non-blind}) fails to accurately approximate the real-world degradation due to the inaccessibility of the reference image.

\begin{figure}\tiny
\centering
\includegraphics[scale=0.36]{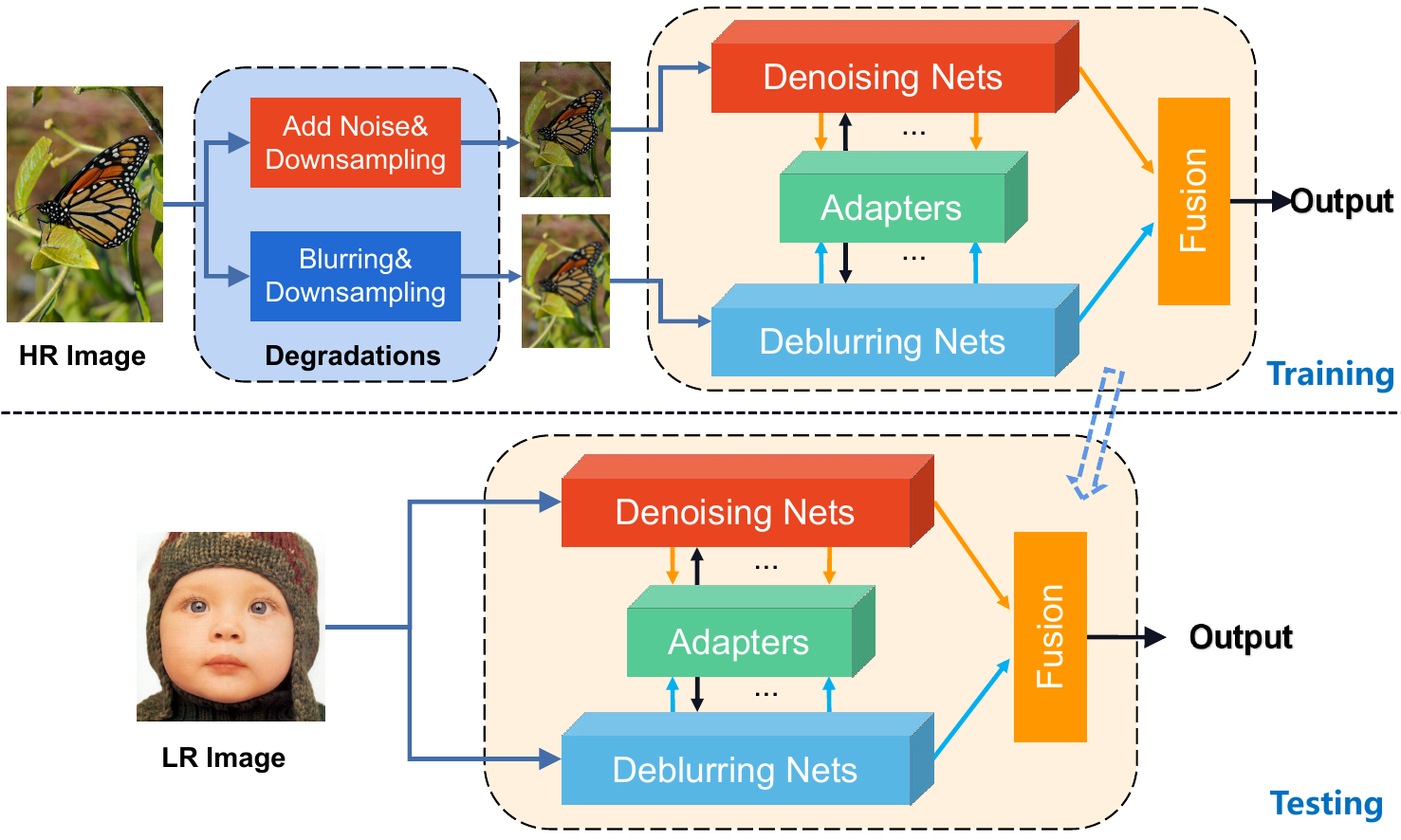}
\vspace{-0.3cm}
\caption{\small The training and testing procedures of RealNet.}
\label{fig:train_test}
\vspace{-0.5cm}
\end{figure}

Subsequently, researchers have developed blind SR methods, which assume that the LR image is obtained from the HR image through a degradation function with a complex and unknown degradation type, and this degradation function is typically assumed to follow the equation:
\begin{equation}
\label{eq:blind}
\mathbf{x}_{lr} = (\mathbf{x}_{hr} \otimes \mathbf{Kernel}) \downarrow_{s} + N,
\end{equation}
where $\otimes$ denotes the convolutional operation, $\mathbf{Kernel}$ represents the unknown blur kernel, $\downarrow_{s}$ signifies the downsampling operation with a scale factor of $s$, and $N$ accounts for sophisticated noise. According to Eq. (\ref{eq:blind}), we can clearly see that a single HR image corresponds to infinitely many possible LR images, which is typically an ill-posed problem. Nonetheless, a multitude of blind SR methods has been proposed to overcome this problem. They can be broadly divided into two main types: methods without blur kernel estimation \cite{SRMD, zhang2021plug, usrnet, manet} and methods with blur kernel estimation \cite{IKC, DANv1, DANv2}.

As for methods without blur kernel estimation, SRMD \cite{SRMD} is the first to combine the preset blur kernel, noise, and LR image as the input to the SR network. After that, \cite{zhang2019deep, usrnet} incorporated Maximum A Posteriori Probability into iterative optimization. For these methods, the selection of parameters for the preset blur kernel is crucial for achieving effective results in practical applications. However, it is challenging to manually choose appropriate preset blur kernel parameters in the real world. To overcome this challenge, researchers have developed some methods to estimate the blur kernel. IKC \cite{IKC} proposed to iteratively estimate the blur kernel, which divides the SR reconstruction process into two steps: 1) estimating the blur kernel from the LR image and 2) combining the estimated blur kernel and the LR image to reconstruct the SR image. However, the independence between the training of the blur kernel estimation model and the SR model leads to incompatibility issues. To address this issue, DANv1 \cite{DANv1} proposed an alternating optimization strategy between the blur kernel estimation and SR models, which makes the two models form a closed loop that can be trained in an end-to-end manner. Although DANv1 has achieved impressive performance in blind SR tasks, NLCUNet \cite{NLCUNet} has made significant advancements by incorporating the Non-Local Sparse Attention (NLSA) \cite{NLSA} module. It screens global features in the feature map by calculating self-similarities, enabling implicit estimation of the blur kernel. Therefore, the self-similarity technique has received the widest application in blind SR tasks. However, blind SR models that incorporate the calculation of self-similarity suffer from high computational complexity, making them unsuitable for resource-constrained devices. It is worth noting that the commonly used module for calculating self-similarity is Non-Local Attention (NLA) \cite{NLA}.

In this paper, we go deep into the execution process of NLA. We find that during the self-similarity calculation, a matrix multiplication operation between the high-dimensional attention map and the Value is required, leading to high computational complexity. The attention map is obtained by multiplying the Query matrix with the Key matrix. Then the softmax function is further performed, which hinders the separation of Query and Key. Based on this finding, we propose a second-order Taylor expansion approximation (STEA) to significantly reduce the computational complexity. Compared to self-similarity calculations, STEA loses the ability to filter global features. Hence, we design a multi-scale large field reception (MLFR) to compensate for the performance loss caused by STEA. Finally, we apply our two core designs to laboratory and real-world scenarios by constructing LabNet and RealNet, respectively. Extensive experiments validate the effectiveness of the proposed method. The major contributions of this paper are summarized as follows.
\begin{itemize}
    \item We propose the second-order Taylor expansion approximation (STEA) strategy, which significantly reduces the computational complexity in self-similarity-based blind SR tasks.
    \item We design a multi-scale large field reception (MLFR) to compensate for the performance loss caused by STEA, showcasing competitive performance over the state-of-the-art.
    \item We construct the LabNet and the RealNet by integrating our two core designs (STEA and MLFR) to respectively verify the effectiveness of our designs in the laboratory and real-world scenarios. Extensive experimental results demonstrate that our LabNet achieves state-of-the-art performance while maintaining linear computational complexity, and our RealNet effectively handles various real-world degradations and achieves superior performance in diverse blind SR scenarios.
\end{itemize}

\begin{figure*}\tiny
\centering
\includegraphics[scale=0.53]{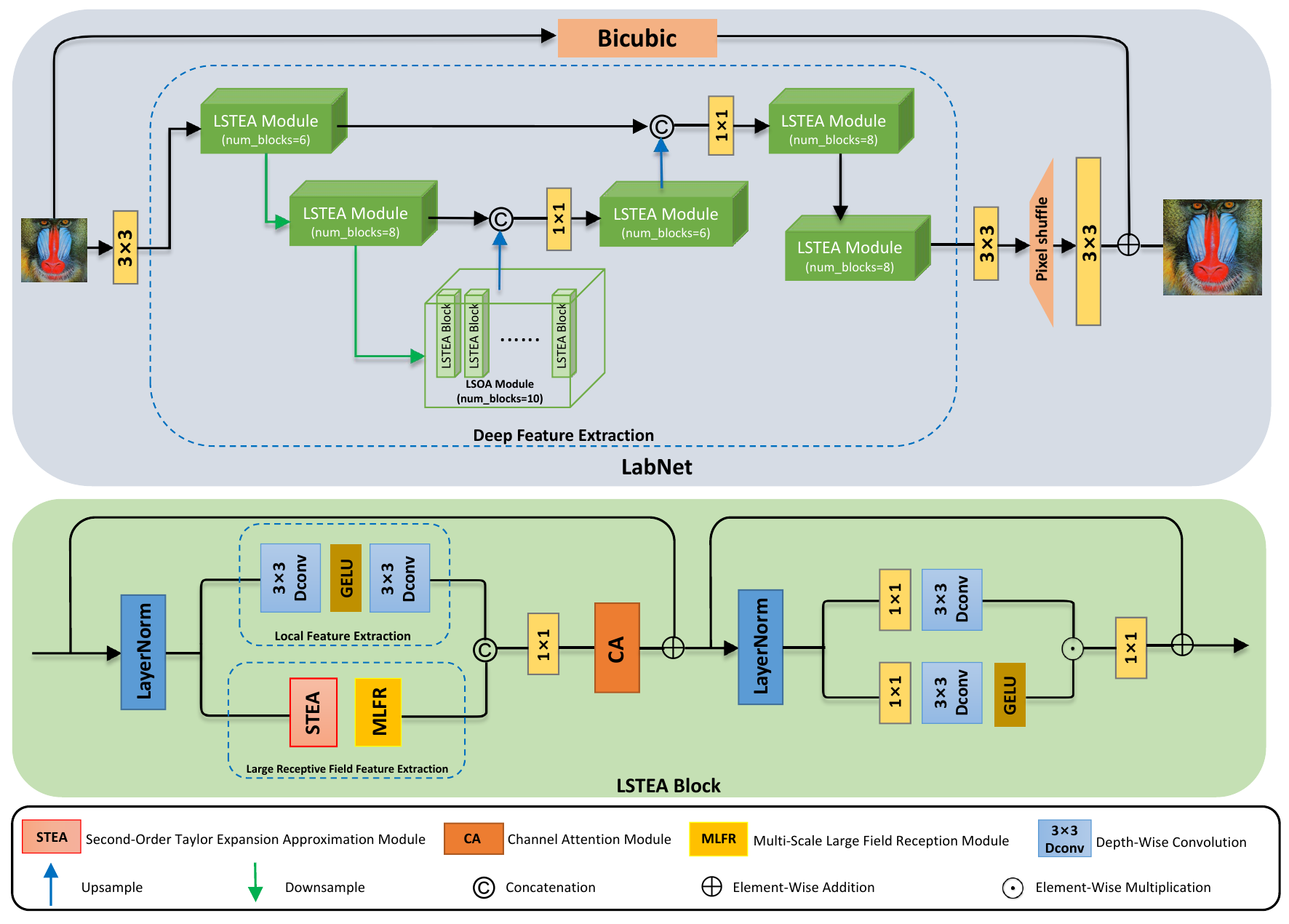}
\vspace{-0.3cm}
\caption{\small The overview of our proposed LabNet.}
\label{fig:LabNet}
\vspace{-0.5cm}
\end{figure*}

The structure of this paper is organized as follows: Section II provides a comprehensive overview of the proposed methodology. Specifically, Section II-A presents a detailed explanation of the intricate aspects of STEA. Section II-B elucidates the specific details of MLFR. Section II-C outlines the implementation specifics of LabNet, while Section II-D delves into the implementation details of RealNet. Subsequently, Section III encompasses the dataset and evaluation metrics (Section III-A), implementation details of LabNet (Section III-B), implementation details of RealNet (Section III-C), comparison with state-of-the-art techniques (Section III-D), ablation studies and analysis of LabNet (Section III-E), ablation study and analysis of RealNet (Section III-F) and adjustable parameter details in RealNet (Section III-G). Section IV concludes the paper.

\section{METHODOLOGY}
In this section, we provide a detailed explanation of the STEA and MLFR algorithms, as well as their deployment to the LabNet (testing for laboratory settings) and RealNet (testing for real-world scenarios).

\begin{figure}[t]\tiny
\centering
\includegraphics[scale=0.23]{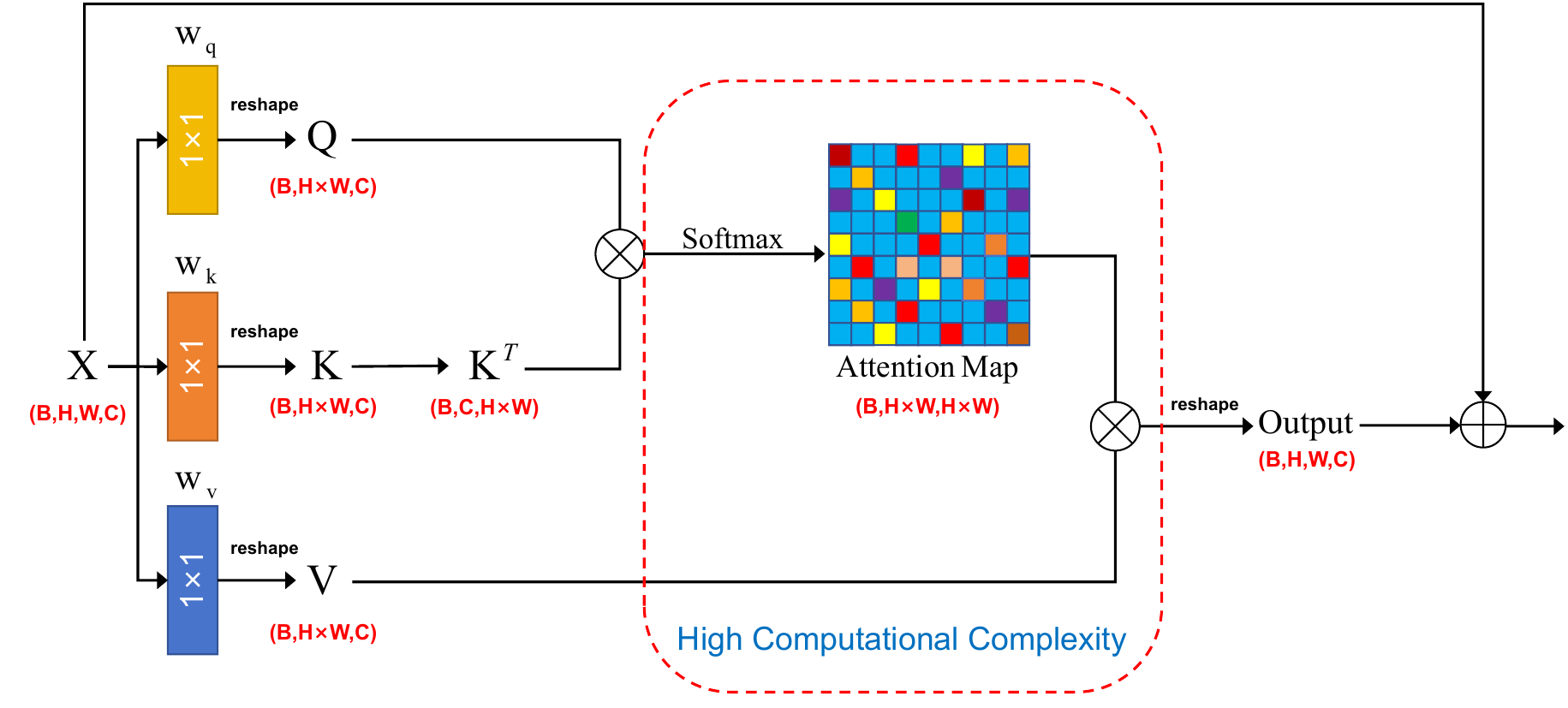}
\vspace{-0.3cm}
\caption{\small Analysis of the execution process of the Non-Local Attention (NLA) module.}
\label{fig:NLA_anl}
\vspace{-0.4cm}
\end{figure}

\subsection{\textbf{Second-Order Taylor Expansion Approximation (STEA)}}
The self-similarity has been widely applied to the blind SR task, which is commonly implemented by a Non-Local Attention (NLA) \cite{NLA}. However, we found that the NLA brings extremely high computational complexity. This mainly arises from the matrix multiplication between the high-dimensional Attention Map and the $V$ matrix, as highlighted by the red dashed box of Fig.~\ref{fig:NLA_anl}. Without loss of generality, $Q$, $K$, $V$ $\in \mathcal{R}^{N\times d}$ denote query, key, and value matrix, respectively. Then, we know the computational complexity of NLA is $\mathcal{O}(N^2\times d)$, where $N$ and $d$ represent the dimension of the feature map and the number of channels, respectively. In this part, we propose an STEA algorithm to significantly reduce the computational complexity of NLA. To this end, we first analyze the execution process of the NLA module. As shown in Fig.~\ref{fig:NLA_anl}, the Attention Map is derived from the matrix multiplication of $Q$ and $K^T$. Due to the limitation imposed by the softmax operation, the matrix multiplication between $Q$ and $K^T$ cannot be separated, resulting in a $\mathcal{O}(N^2)$ complexity. To address this challenge, we propose a Second-Order Taylor Expansion to approximate the NLA module. Here, we mathematically rewrite the execution process of the NLA as follows:
\begin{equation}\label{QKV}
    Q = XW_{q}, K = XW_{k}, V = XW_{v},
\end{equation}
where $W_{q}$, $W_{k}$ and $W_{v}$ stand for the parameters of three $1\times 1$ convolutions, respectively. $Q$, $K$ and $V$ represent the Query, Key and Value, respectively. $X$ denotes the input feature map. Next, we perform the matrix multiplication between $Q$ and $K^T$, followed by the softmax operation, to obtain the Attention Map:
\begin{equation}\label{Attmap}
    AttMap = \textrm{SoftMax}(QK^T),
\end{equation}
where $AttMap$ represents the Attention Map, and $\textrm{SoftMax}(\cdot)$ denotes the softmax function. By substituting Eq. (\ref{QKV}) to Eq. (\ref{Attmap}), we have
\begin{equation}
AttMap = \textrm{SoftMax}(XW_{q}W_{k}^TX^T). 
\end{equation}
We denote $W_{qk} = W_{q}W_{k}^T$, resulting in the below expression:
\begin{equation}\label{eq:AttMap}
    AttMap = \textrm{SoftMax}(XW_{qk}X^T).
\end{equation}
In the original NLA module, the multiplication between $Q$ and $K^T$ followed by the softmax operation gives rise to softmax attention. Although softmax attention can effectively enhance the performance of the SR network by making the matrix full-rank, it introduces high computational complexity since such a high-dimensional Attention Map ($AttMap$) must be multiplied with $V$ in the subsequent computation. For that, we propose to simplify the softmax function as an exponential form:
\begin{equation}
\label{eq:Taylor}
   AttMap = \textrm{SoftMax}(QK^T) = \frac{1}{k} \times e^{QK^T} = \frac{1}{k} \times e^{XW_{qk}X^T},
\end{equation}
where $k$ is a learnable parameter used to approximate the denominator of softmax. By now, we can see that the operation between $QK^T$ and $V$ is still nonlinear due to the presence of an exponential function, which disables us from separating $Q$ and $K$ from $QK^T$. In subsequent multiplication by $V$, the basic operation of the matrix cannot be flexibly used, failing to reduce the high computational complexity. For this purpose, we exploit a second-order Taylor expansion as a substitute for the exponential function, achieving a good trade-off between

\begin{figure}[t]\tiny
\centering
\includegraphics[scale=0.15]{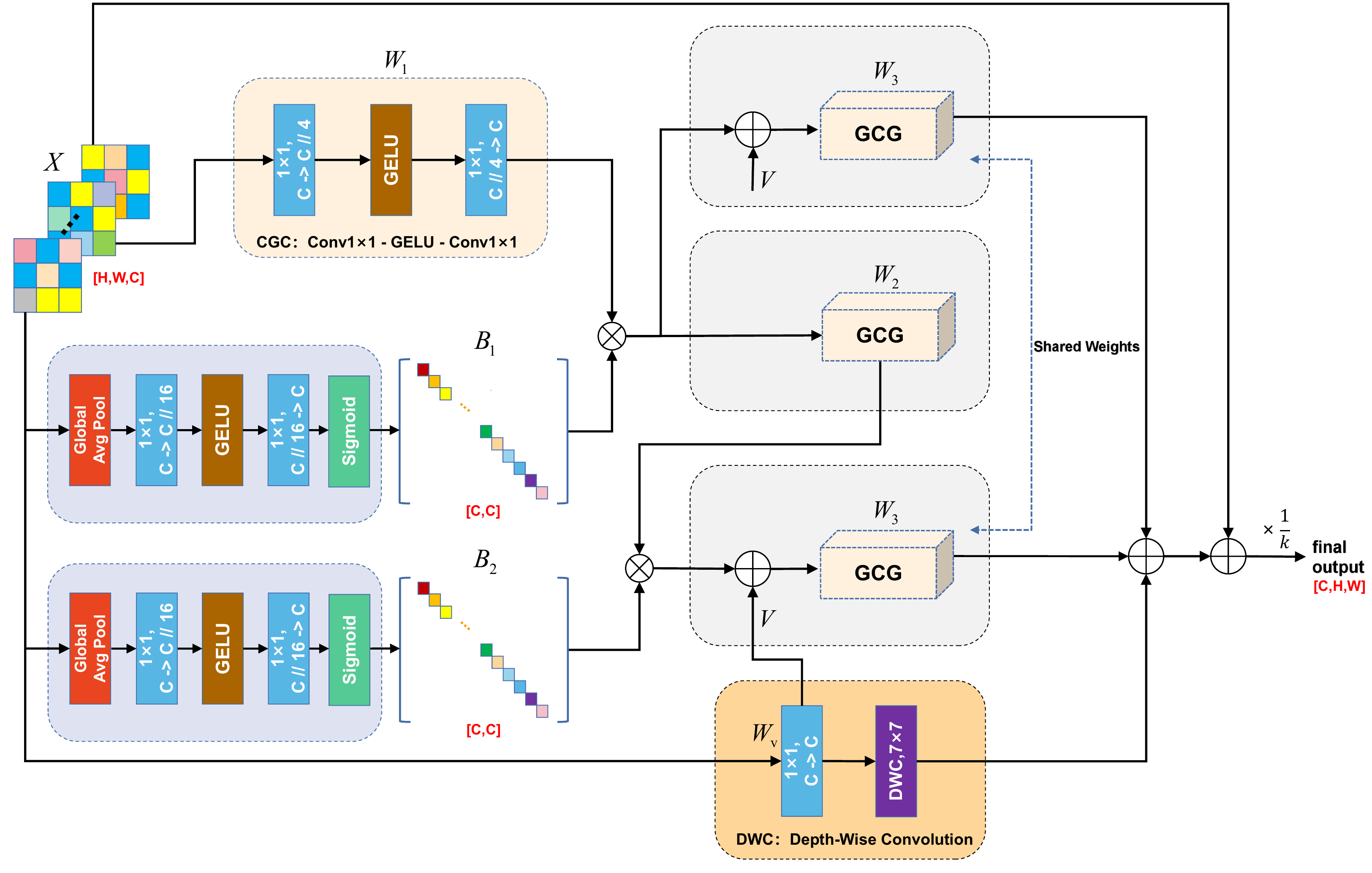}
\vspace{-0.3cm}
\caption{\small The Second-Order Taylor Expansion Approximation (STEA) module.}
\label{fig:STEA}
\vspace{-0.4cm}
\end{figure}

\noindent
efficiency and accuracy. 

We know that in principle, the more remainder terms kept in the Taylor expansion of the exponential function, the more the expansion approximates the raw function. According to our empirical results and considering a good trade-off between accuracy and complexity, we adopt the second-order Taylor expansion on Eq. (\ref{eq:Taylor}) as follows:
\begin{multline}
\label{eq:feamap}
   AttMap = \frac{1}{k} \times e^{XW_{qk}X^T} \\ = \frac{1}{k} \times (I_n + XW_{qk}X^T + \frac{(XW_{qk}X^T)^2}{2!} + O(XW_{qk}X^T)),
\end{multline}
where $I_n$ stands for identity matrix and $O(\cdot)$ stands for the high-order remainder of the Taylor expansion. By ignoring the constant and remainder terms of Eq.(\ref{eq:feamap}), we achieve:
\begin{multline}
\label{eq:feamap2}
AttMap= \frac{1}{k} \times (I_n + XW_{qk}X^T + \\ XW_{qk}X^TXW_{qk}X^T).
\end{multline}
Then, we multiply $AttMap$ by $V$ and refer to Eq.(\ref{QKV}) to obtain:
\begin{multline}
\label{eq:SoftV}
   Output = AttMap\times V = \frac{1}{k} \times (XW_{v} + \\ XW_{qk}X^TXW_{v} + XW_{qk}X^TXW_{qk}X^TXW_{v}).
\end{multline}
Eq. (\ref{eq:SoftV}) clearly indicates that the constraint of the softmax function has been removed, and therefore we can apply the associative property of the matrix operation in Eq. (\ref{eq:SoftV}), which allows us to calculate $X^TX$ first. Considering that $X^TX$ is a symmetric matrix, we can transform $X^TX$ into a diagonalization expression as follows:  
\begin{equation}
\label{eq:diagonal}
    X^TX = ZBZ^T,
\end{equation}
where $B$ is a diagonal matrix with its diagonal elements being the eigenvalues of $X^TX$, and $Z$ is an orthogonal matrix whose columns are the eigenvectors of $X^TX$ corresponding to the eigenvalues. By substituting Eq. (\ref{eq:diagonal}) into Eq. (\ref{eq:SoftV}), we obtain:
\begin{multline}
\label{eq:diagonal12}
    Output = \frac{1}{k} \times (XW_{v} + XW_{qk}ZBZ^TW_{v} + \\ XW_{qk}ZBZ^TW_{qk}ZBZ^TW_{v}).
\end{multline}
According to Eq. (\ref{eq:diagonal}), we can see that obtaining $B$ through the matrix eigen-decomposition operation is computationally expensive. If we randomly initialize a learnable diagonal matrix $B$ to avoid the high computational complexity from the eigen-decomposition operation of $X^TX$. Unfortunately, this strategy makes the matrix $B$ independent of the input feature maps ($X$). Besides, in the random initialization, the matrix $B$ is required to be a fixed size, making $X$ have to be the fixed size. This disables the input LR image to be an arbitrary size. To reduce the complexity of the matrix eigen-decomposition and implement the SR task on the LR images with an arbitrary size, we design two separative eigenvalue extraction modules to respectively obtain two diagonal matrices $B_1$ and $B_2$, as shown in Fig.~\ref{fig:STEA}. Our design avoids randomly initializing a single $B$, and meanwhile enhances the representation ability of the model. Therefore, we can obtain a learning-based approximation expression as follows: 
\begin{multline}
    Output = \frac{1}{k} \times (XW_{v} + XW_{qk}ZB_{1}Z^TW_{v} + \\ XW_{qk}ZB_{1}Z^TW_{qk}ZB_{2}Z^TW_{v}).
\end{multline}
It is observed from Fig.~\ref{fig:STEA} that our design modules can dynamically extract global information from the matrices, which can well match the matrix eigen-decomposition where global information is exploited dynamically. 

Similarly, we adopt learnable parameters to substitute $Z$, solving the high computational complexity of the eigen-decomposition of $X^TX$. Next, we let $W_{1} = W_{qk}Z$, $W_{2} = Z^TW_{qk}Z$, and $W_{3} = Z^TW_{v}$. Hence, we obtain:
\begin{equation}
\label{eq:final}
    Output = \frac{1}{k} \times (XW_{v} + XW_{1}B_{1}W_{3} + XW_{1}B_{1}W_{2}B_{2}W_{3}).
\end{equation}
By now, we equate the softmax function to an exponential function and perform a Taylor expansion, making the operations of $Q$, $K$, and $V$ more flexible. It is worth noting that the softmax function can transform the resulting matrix derived from $Q$ and $K^T$ into a full-rank matrix, while the matrix obtained by direct multiplication of $Q$ and $K^T$ is not a full-rank matrix, which significantly reduces the feature diversity, in turn, leads to a degradation of the model's performance \citep{FLatten}. Therefore, as done in \citep{FLatten}, we perform a depthwise convolution (DWC) operation on $XW_{v}$ of Eq. (\ref{eq:final}) to compensate for the performance degradation. Finally, we obtain:
\begin{multline}
\label{eq:final_final}
    FinalOutput = \frac{1}{k} \times [\textrm{DWC}(XW_{v}) + \\ XW_{1}B_{1}W_{3} + XW_{1}B_{1}W_{2}B_{2}W_{3}].
\end{multline}
As shown in Fig.~\ref{fig:STEA}, we design the same network block to learn $W_1$, $W_2$, and $W_3$, which first reduces the number of channels to a quarter and then restores it to the original channel number. Meanwhile, an efficient $Conv_{1\times1}$ operation is adopted to approximate $W_v$. 

As shown in Fig.~\ref{fig:NLA_anl}, $N$ and $d$ are a multiplication of the width and height of a feature map and the number of channels. In general, $N$ is far larger than $d$. According to Eq. (\ref{eq:final_final}), our STEA strategy reduces the complexity of the original NLA from $\mathcal{O}(N^2\times d)$ to $\mathcal{O}(N\times d^2)$. The subsequent experimental results validate that our STEA is more efficient compared to the NLA module, however, exhibits a slight performance degradation due to the ignoring of remaining terms of Taylor extension and the use of the learning-based matrix. To compensate for such performance degradation, we design a Multi-Scale Large Field Reception (MLFR), which will be detailedly described in the following.

\begin{figure}[t]\tiny
\centering
\includegraphics[scale=0.23]{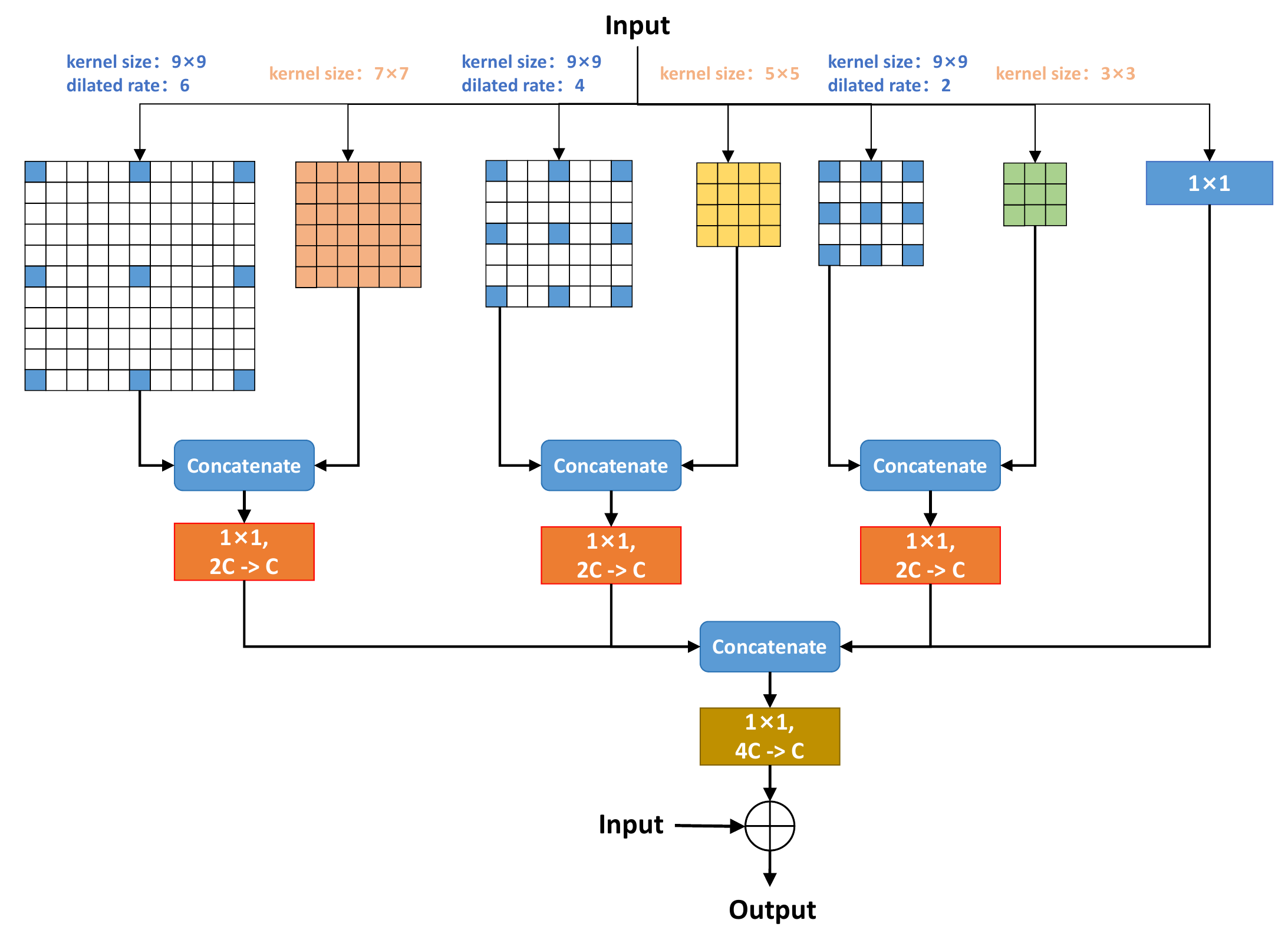}
\vspace{-0.3cm}
\caption{\small The Multi-Scale Large Field Reception (MLFR) module.}
\label{fig:MLFR}
\vspace{-0.4cm}
\end{figure}

\subsection{\textbf{Multi-Scale Large Field Reception (MLFR)}}
As mentioned in Section II-A, compared to the NLA module, our STEA achieves higher efficiency but sacrifices

\begin{figure*}\tiny
\centering
\includegraphics[scale=0.46]{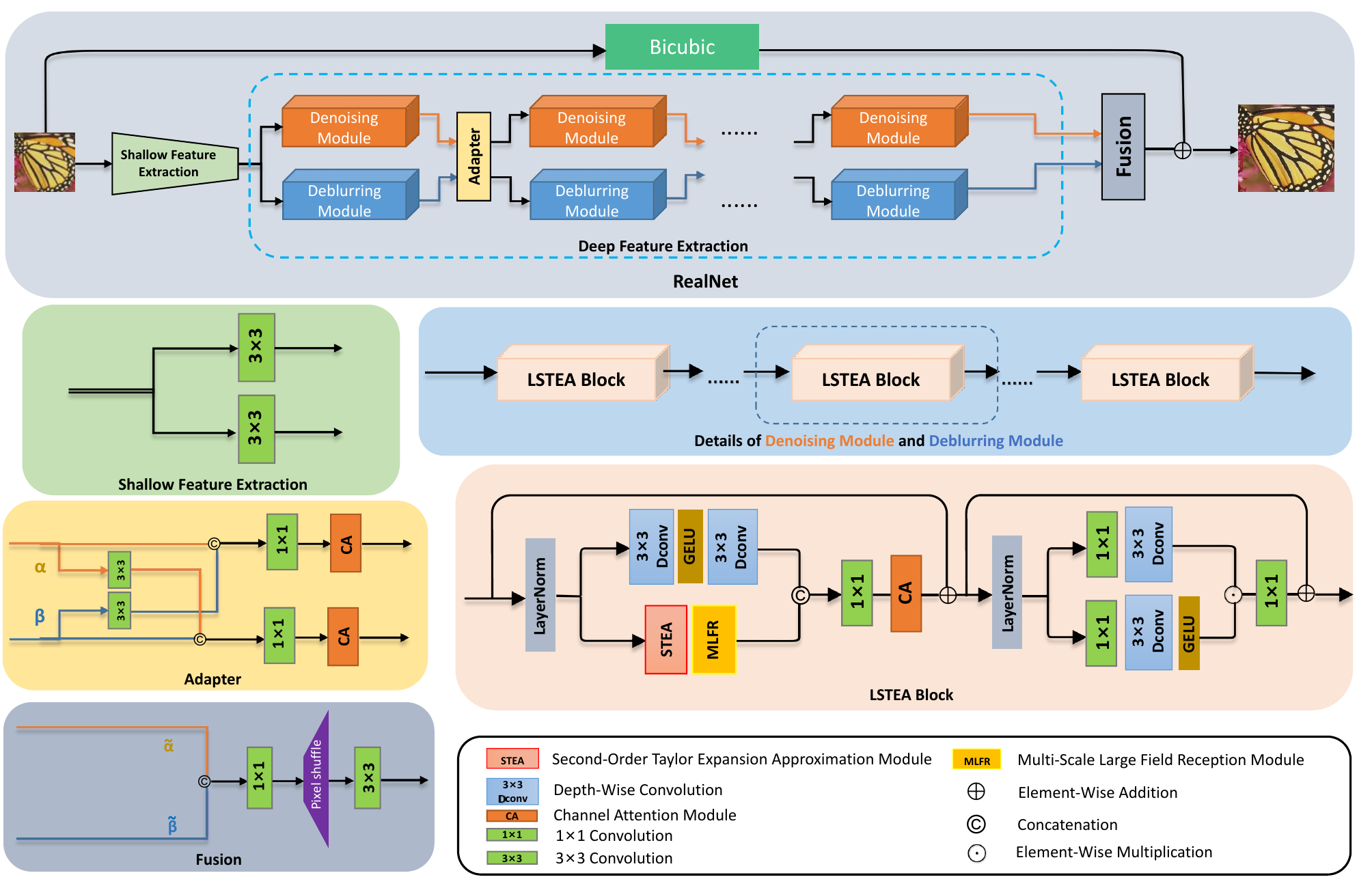}
\vspace{-0.3cm}
\caption{\small The overview of our proposed RealNet.}
\label{fig:RealNet}
\vspace{-0.4cm}
\end{figure*}

\noindent
the ability to extract global features from the feature maps. To this end, we design a Multi-Scale Large Field Reception (MLFR) strategy to obtain a large receptive field, as shown in Fig.~\ref{fig:MLFR}. For the unfilled parts of the dilated convolution, we use normal convolution to fill in the dilation. Specifically, we process the input feature map $X$ through three separate branches. In the first branch, the input is processed by a dilated convolution with an expansion rate of 6 and a depth-wise convolution with a kernel size of ${7\times7}$, respectively. The output results are concatenated along the channel dimension. Simultaneously, in the second branch, the input is processed by a dilated convolution with an expansion rate of 4 and a depth-wise convolution with a kernel size of ${5\times5}$, respectively. Similarly, the output results are concatenated along the channel dimension. In the last branch, the input is processed by a dilated convolution with an expansion rate of 2 and a depth-wise convolution with a kernel size of ${3\times3}$, respectively. Again, the output results are concatenated along the channel dimension. We represent the above execution process using the following equation: 
\begin{multline}
\label{eq:3_branchs}
Output_{(i,j)}=\textrm{Concat}(\textrm{DDWC}(X)_{dr=i}, \\ \textrm{DWC}(X)_{ks=j})_{{\substack{i=2,4,6\\j=3,5,7}}},
\end{multline}
where Concat($\cdot,\cdot$) stands for concatenate operation, $\textrm{DDWC}(\cdot)$ stands for dilation convolution operation on the input, $dr$ is the dilation rate, and $\textrm{DWC}(\cdot)$ stands for depth-wise convolution operation on the input, $ks$ is the convolution kernel size. Furthermore, each value of $i$ corresponds uniquely to one value of $j$. Subsequently, we employ ${1\times1}$ convolutions to perform channel-wise feature fusion on the outputs of the three branches, effectively reducing the dimension of the channels:
\begin{equation}
\label{eq:3_branchs_fuse}
    Fusion_{(i,j)} = \textrm{Conv}_{1\times1}(Output_{(i,j)})_{\substack{i=2,4,6\\j=3,5,7}},
\end{equation}
where $Fusion_{(i,j)}$ represents the output obtained after channel-wise feature fusion of the input, and $\textrm{Conv}_{1\times1}(\cdot)$ denotes the ${1\times1}$ operation performed on the input. Similarly, each value of $i$ corresponds uniquely to one value of $j$. In addition, the input $X$ undergoes a ${1\times1}$ bypass convolution in addition to the aforementioned three branches:
\begin{equation}
\label{eq:bypass}
    Bypass = \textrm{Conv}_{1\times1}(X).
\end{equation}
The purpose of the ${1\times1}$ convolution in Eq. (\ref{eq:bypass}) is to perform subtle adjustments to the channels of the input $X$. Finally, we concatenate the outputs of the three branches with the output of the bypass ${1\times1}$ convolution along the channel dimension and perform channel-wise feature fusion using a ${1\times1}$ convolution to reduce the channel dimension:
\begin{multline}
\label{eq:final_MLFR}
FinalOutput=Conv_{1\times1}(Concat(Fusion_{(i,j)}, \\ Bypass)_{\substack{i=2,4,6\\j=3,5,7}}).
\end{multline}
In summary, we utilize multiple dilated convolutions with different scales to extract features with various receptive fields. For the unpadded regions in dilated convolutions, we employ depth-wise convolutions to fill in the dilation. Additionally, we utilize ${1\times1}$ convolutions to perform subtle adjustments on the input as bypass convolutions. In the final step of feature fusion, we concatenate all the branches along the channel dimension and then employ a ${1\times1}$ convolution for channel-wise feature fusion, resulting in the final output as shown in Eq. (\ref{eq:final_MLFR}).

\begin{figure}\tiny
\centering
\includegraphics[scale=0.24]{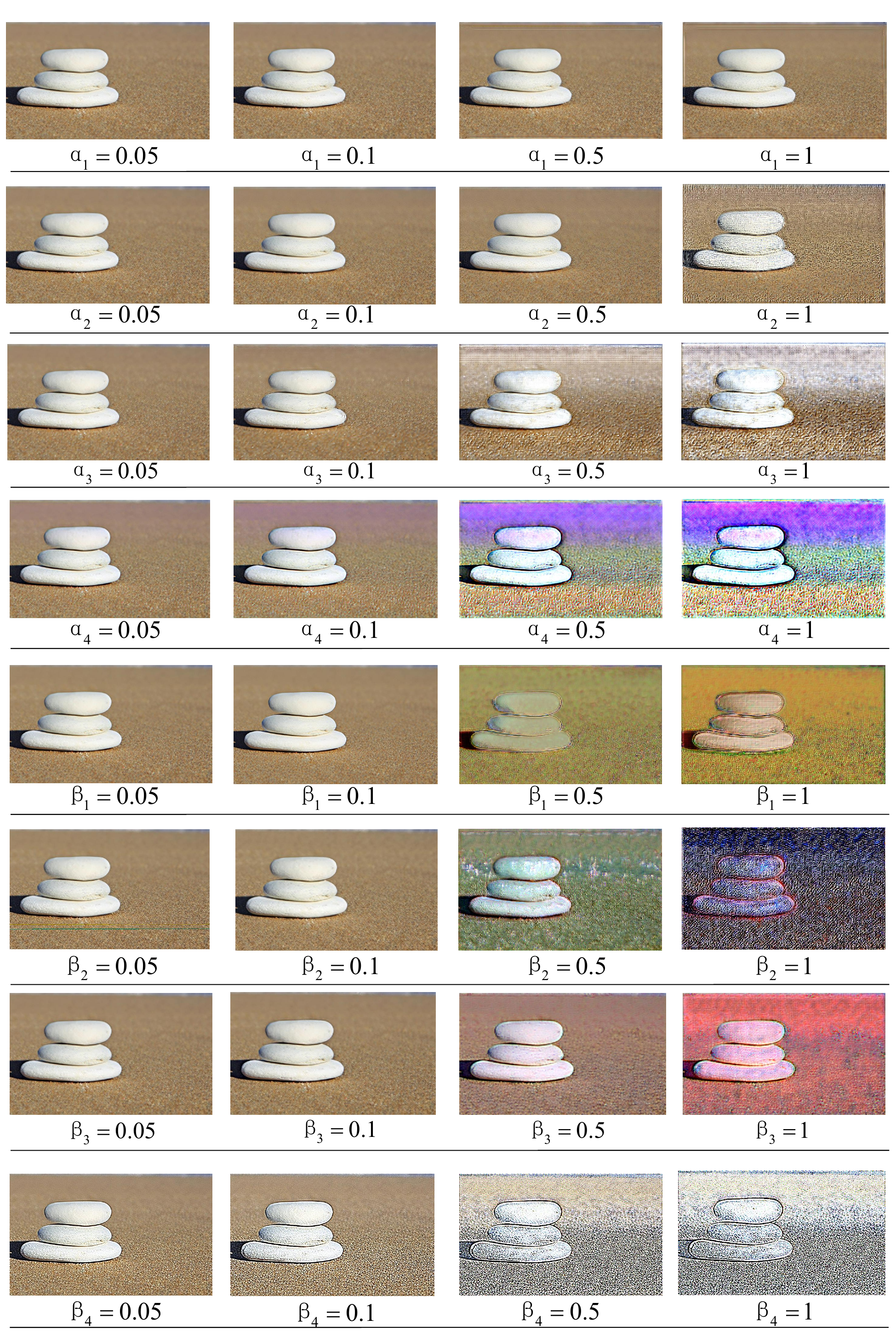}
\vspace{-0.4cm}
\caption{\small Users can adjust the eight parameters in the RealNet according to their needs. The above figure shows the effect obtained by adjusting different parameters. When adjusting one of the parameters, we set the other parameters to 0.01.}
\label{fig:single_params}
\vspace{-0.4cm}
\end{figure}

\subsection{\textbf{Application to Laboratory Scenario (LabNet)}}
To show the effectiveness of our STEA and MLFR designs in a laboratory scenario, we construct a novel network termed LabNet for testing. As shown in Fig.~\ref{fig:LabNet}, our LabNet is an end-to-end blind SR framework composed of three parts: shallow feature extraction, deep feature extraction, and HR image reconstruction. For the shallow feature extraction part, we employ a ${3\times3}$ convolution to extract shallow features from the input LR image. Subsequently, the extracted shallow features are fed into the UNet for deep feature extraction. Each LSTEA module in the UNet consists of multiple LSTEA blocks. When the input features pass through an LSTEA block, they undergo a channel-wise LayerNorm operation, followed by two branches. One branch utilizes multiple depth-wise convolutions for local feature extraction, while the other branch employs the STEA and MLFR modules for large-scale feature extraction. The channel features are then fused using a ${1\times1}$ convolution and filtered using the channel attention module. Subsequently, the features pass through the GDFN \cite{GDFN} module, which performs a nonlinear mapping operation on the input to enhance network performance. Finally, for the HR image reconstruction part, the channel dimension is adjusted using a ${3\times3}$ convolution. The $H$ and $W$ dimensions are then expanded to the target size using a pixel shuffle operation, and a final refinement is performed using a ${3\times3}$ convolution to obtain the ultimate output results.

\subsection{\textbf{Application to Real-world Scenario (RealNet)}}
To show the effectiveness of our STEA and MLFR designs in real-world scenarios, we construct a novel network termed RealNet for testing. As illustrated in Fig.~\ref{fig:RealNet}, our RealNet comprises three key components: shallow feature extraction, deep feature extraction, and HR image reconstruction. Firstly, we perform shallow feature extraction on the input LR image using two separate ${3\times3}$ convolutions. Subsequently, the extracted features are separately subjected to deep feature extraction in the denoising branch and the deblurring branch. Both the Denoising and Deblurring modules are composed of multiple LSTEA blocks. The features extracted from the Denoising module and the Deblurring module are then fused in proportion using Adapter modules. Finally, we employ a Fusion module for HR image reconstruction.\\
$\textbf{Adapter Module.}$ For images contaminated by noise, there are numerous detailed features within the image that can be learned by the deblurring branch. Similarly, for blurry images, there are many low-frequency features within the image that can be learned by the denoising branch. Therefore, as depicted in Fig.~\ref{fig:RealNet}, we propose the Adapter module. The parameters $\alpha$ and $\beta$ control the proportions of denoising and deblurring, respectively. Analyzing the denoising branch, we take the features extracted by the deblurring branch in the previous step, apply a ${3\times3}$ convolution, and perform a channel-wise concatenation with the features extracted by the denoising branch. Subsequently, a ${1\times1}$ convolution is employed for channel-wise feature fusion, followed by a channel attention module for channel-wise feature selection. Similarly, a similar processing approach is utilized for the deblurring branch.\\
$\textbf{Fusion Module.}$ To adjust the proportions of the outputs from the final denoising module and deblurring module, we also utilize two adjustable parameters ($\alpha$ and $\beta$). After adjusting the two sets of output feature maps, we perform a channel-wise concatenation and then apply a 1$\times$1 convolution for channel information fusion and adjustment of channel dimensions. Finally, the output feature maps are resized to match the target size using a pixel shuffle operation and fine-tuned with a ${3\times3}$ convolution.\\
$\textbf{RealNet-GAN.}$ To achieve better visual results, we employ a GAN training strategy for the proposed RealNet. Specifically, we use RealNet as the generator and utilize a U-Net as the discriminator \cite{Real-ESRGAN}. The training process consists of two parts. Firstly, we train a PSNR-oriented model using the $\ell_1$ loss, which is then used as the generator in the GAN training strategy. We then train a GAN-oriented model using a combination of $\ell_1$ loss \cite{L1_loss}, perceptual loss \cite{Perceptual}, and adversarial loss \cite{adversarial}. The overall loss function of our network is
\begin{equation}
    \mathcal{L}_{entire} = \mathcal{L}_1 + \mathcal{L}_{perc} + 0.1 \times \mathcal{L}_{adv},
\end{equation}
where $\mathcal{L}_1$, $\mathcal{L}_{perc}$, and $\mathcal{L}_{adv}$ represent the $\ell_1$ loss, perceptual loss, and adversarial loss, respectively. According to our empirical study, their weights are set to be $1$, $1$, and $0.1$, respectively. 
We utilize the \{conv1,..., conv5\} feature maps before activation in the pre-trained VGG19 \cite{vgg} model (with weights \{$0.1$, $0.1$, $1$, $1$, $1$\}) as the perceptual loss.

\begin{table}[t]
\centering
\caption{Comparison of inference times and GFLOPs with different Blind SR methods.}
\label{tab:sota_GFLOPs}
\resizebox{\columnwidth}{!}{%
\begin{tabular}{cccc}
\hline
(1,3,64,64) $\rightarrow$ (1,3,256,256) & Infer\_time on GPU & GFLOPs \\ \hline
DCLS                                   & 0.86 s              & 46     \\
NLCUNet                                & 2.23 s              & 432     \\
LabNet (Ours)                   & 1.18 s             & 49      \\ \hline 
\end{tabular}%
}
\end{table}

\section{Experiments}
\label{sec:experiment}

\subsection{\textbf{Datasets and Evaluation Metrics}}
We train all our proposed SR models using the \textbf{DF2K} dataset (\textbf{DIV2K} \cite{DIV2K} + \textbf{Flickr2K} \cite{Flickr2K}). There are 3450 2K-resolution images in the dataset. For the quantitative evaluation, we conducted tests on five standard super-resolution image benchmark datasets: Set5 \cite{Set5}, Set14 \cite{Set14}, BSD100 \cite{BSD100}, Urban100 \cite{Urban100}, and Manga109 \cite{Manga109}. To conduct a qualitative evaluation, we compared the visual results of our approach with previous methods on the RealWorld38 \cite{Real-ESRGAN} dataset. The quantitative performance metrics we use for comparison are the Peak Signal-to-Noise Ratio (PSNR) \cite{psnr} and Structural Similarity (SSIM) \cite{ssim}.

\begin{figure*}[t]\tiny
\centering
\includegraphics[scale=0.6]{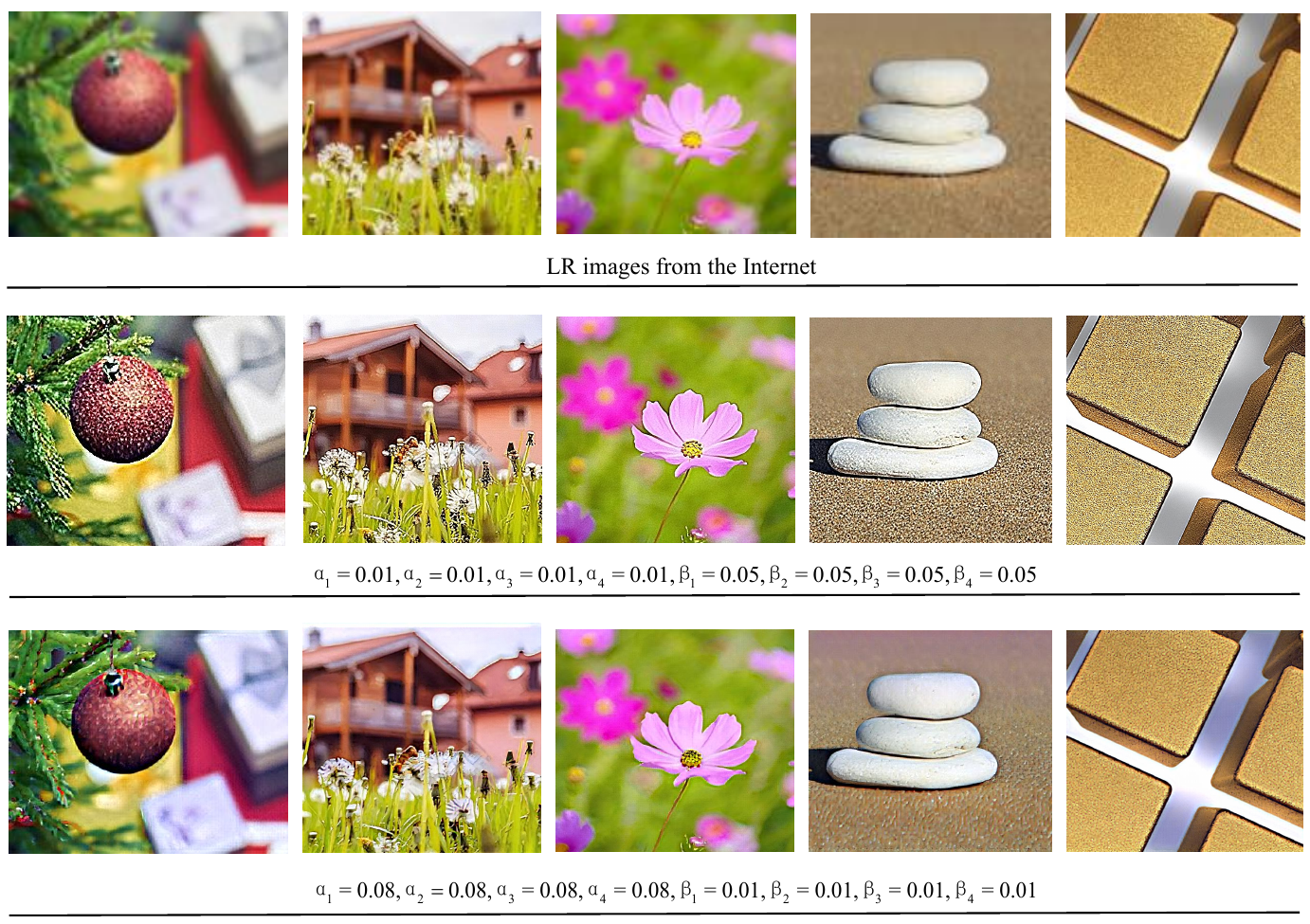}
\caption{\small The images in the above figure were randomly obtained by us from the Internet. We can adjust the eight parameters in RealNet at the same time according to the effect we need.}
\label{fig:multi_params}
\end{figure*}

\begin{table*}[t]\tiny
\centering
\caption{Quantitative results on SISR benchmark datasets. The best and the second best results are \textbf{highlighted} and \underline{underlined}. All methods under the dotted line were tested using the cropped datasets.}
\label{tab:sota_compare}
\resizebox{\textwidth}{!}{%
\begin{tabular}{cccccccccccc}
\hline
Method          & Scale & \multicolumn{2}{c}{Set5}                                     & \multicolumn{2}{c}{Set14}                                    & \multicolumn{2}{c}{BSD100}                                   & \multicolumn{2}{c}{Urban100}                                 & \multicolumn{2}{c}{Manga109}                                 \\
                &       & PSNR                         & SSIM                          & PSNR                         & SSIM                          & PSNR                         & SSIM                          & PSNR                         & SSIM                          & PSNR                         & SSIM                          \\ \hline
Bicubic         &       & 28.82                        & 0.8577                        & 26.02                        & 0.7634                        & 25.92                        & 0.7310                        & 23.14                        & 0.7258                        & 25.60                        & 0.8498                        \\
CARN \citep{CARN}           &       & 30.99                        & 0.8779                        & 28.10                        & 0.7879                        & 26.78                        & 0.7286                        & 25.27                        & 0.7630                        & 26.86                        & 0.8606                        \\
Bicubic+ZSSR \citep{ZSSR}    &       & 31.08                        & 0.8786                        & 28.35                        & 0.7933                        & 27.92                        & 0.7632                        & 25.25                        & 0.7618                        & 28.05                        & 0.8769                        \\
Deblurring \citep{Deblurring}+CARN \citep{ZSSR} &       & 24.20                        & 0.7496                        & 21.12                        & 0.6170                        & 22.69                        & 0.6471                        & 18.89                        & 0.5895                        & 21.54                        & 0.7946                        \\
CARN \citep{ZSSR}+Deblurring \citep{Deblurring} &   & 31.27                        & 0.8974                        & 29.03                        & 0.8267                        & 28.72                        & 0.8033                        & 25.62                        & 0.7981                        & 29.58                        & 0.9134                        \\
IKC \citep{IKC}            & ×2       & 37.19                        & 0.9526                        & 32.94                        & 0.9024                        & 31.51                        & 0.8790                        & 29.85                        & 0.8928                        & 36.93                        & 0.9667                        \\
\hdashline
DANv1 \citep{DANv1}          &       & 37.24                        & 0.9551                        & 33.07                        & 0.9042                        & 31.75                        & 0.8857                        & 30.59                        & 0.9059                        & 37.20                        & 0.9709                        \\
DANv2 \citep{DANv2}          &       & 37.54                        & 0.9571                        & 33.44                        & 0.9095 & 31.99                        & 0.8904 & 31.43                        & 0.9173                        & 38.05                        & 0.9733                        \\
DCLS \citep{DCLS}           &       & 37.60 & 0.9552 & 33.46 & \underline{0.9105} & 32.03 & \underline{0.8906} & 31.69 & \underline{0.9202} & 38.28 & 0.9739 \\
NLCUNet \citep{NLCUNet}   &       & \textbf{37.71} & \underline{0.9571} & \underline{33.60} & 0.9092                       & \underline{32.09} & 0.8901                        & \underline{31.77} & \underline{0.9202} & \underline{38.60} & \underline{0.9744} \\

BSRGAN \citep{bsrgan}          &       & 32.24                        & 0.9096                        & 29.53                        & 0.8336 & 28.88                        & 0.8021 & 26.70                        & 0.8303                        & 30.78                        & 0.9189                        \\
Real-ESRGAN \citep{Real-ESRGAN}          &       & 30.20                        & 0.8906                        & 27.96                        & 0.7966 & 27.56                        & 0.7639 & 25.37                        & 0.7921                        & 29.18                        & 0.9046                        \\

LabNet (Ours)   &       & \underline{37.68} & \textbf{0.9575} & \textbf{33.63} & \textbf{0.9111}                        & \textbf{32.14} & \textbf{0.8924}                        & \textbf{31.88} & \textbf{0.9220} & \textbf{38.62} & \textbf{0.9748} \\ \hline
Bicubic         &       & 26.21                        & 0.7766                        & 24.01                        & 0.6662                        & 24.25                        & 0.6356                        & 21.39                        & 0.6203                        & 22.98                        & 0.7576                        \\
CARN \citep{CARN}          &       & 27.26                        & 0.7855                        & 25.06                        & 0.6676                        & 25.85                        & 0.6566                        & 22.67                        & 0.6323                        & 23.85                        & 0.7620                        \\
Bicubic+ZSSR \citep{ZSSR}   &       & 28.25                        & 0.7989                        & 26.15                        & 0.6942                        & 26.06                        & 0.6633                        & 23.26                        & 0.6534                        & 25.19                        & 0.7914                        \\
Deblurring \citep{Deblurring}+CARN \citep{ZSSR} &       & 19.05                        & 0.5226                        & 17.61                        & 0.4558                        & 20.51                        & 0.5331                        & 16.72                        & 0.5895                        & 18.38                        & 0.6118                        \\
CARN \citep{ZSSR}+Deblurring \citep{Deblurring} &    & 30.31                        & 0.8562                        & 27.57                        & 0.7531                        & 27.14                        & 0.7152                        & 24.45                        & 0.7241                        & 27.67                        & 0.8592                        \\
IKC \citep{IKC}            & ×3       & 33.06                        & 0.9146                        & 29.38                        & 0.8233                        & 28.53                        & 0.7899                        & 24.43                        & 0.8302                        & 32.43                        & 0.9316                        \\
\hdashline
DANv1 \citep{DANv1}          &       & 33.81                        & 0.9199                        & 30.08                        & 0.8287                        & 28.94                        & 0.7932                        & 27.64                        & 0.8351                        & 33.08                        & 0.9375                        \\
DANv2 \citep{DANv2}          &       & 33.87                        & 0.9208                        & 30.20                        & 0.8311                        & 29.02                        & 0.7962                        & 27.82                        & 0.8395                        & 33.24                        & 0.9398                        \\
DCLS \citep{DCLS}           &        & 33.95 & 0.9214 & 30.28 & 0.8328 & \underline{29.06} & 0.7970 & 28.02 & 0.8443 & 33.48 & 0.9412 \\
NLCUNet \citep{NLCUNet}  &       & \textbf{34.31} & \textbf{0.9231} & \underline{30.37} & \underline{0.8337} & \textbf{29.14} & \underline{0.7983} & \textbf{28.30} & \underline{0.8479} & \underline{33.71} & \textbf{0.9436} \\ 
LabNet (Ours)   &       & \underline{34.21} & \underline{0.9228} & \textbf{30.38} & \textbf{0.8349} & \textbf{29.14} & \textbf{0.7998} & \underline{28.25} & \textbf{0.8494} & \textbf{33.73} & \underline{0.9431} \\ \hline
Bicubic         &       & 24.57                        & 0.7108                        & 22.79                        & 0.6032                        & 23.29                        & 0.5786                        & 20.35                        & 0.5532                        & 21.50                        & 0.6933                        \\
CARN \cite{CARN}           &       & 26.57                        & 0.7420                        & 24.62                        & 0.6226                        & 24.79                        & 0.5963                        & 22.17                        & 0.5865                        & 21.85                        & 0.6834                        \\
Bicubic+ZSSR \cite{ZSSR}   &       & 26.45                        & 0.7279                        & 24.78                        & 0.6268                        & 24.97                        & 0.5989                        & 22.11                        & 0.5865                        & 23.53                        & 0.7240                        \\
Deblurring \cite{Deblurring}+CARN \cite{ZSSR} &       & 18.10                        & 0.4843                        & 16.59                        & 0.3994                        & 18.46                        & 0.4481                        & 15.47                        & 0.3872                        & 16.78                        & 0.5371                        \\
CARN \cite{ZSSR}+Deblurring \cite{Deblurring} &    & 28.69                        & 0.8092                        & 26.40                        & 0.6926                        & 26.10                        & 0.6528                        & 23.46                        & 0.6597                        & 25.84                        & 0.8035                        \\
IKC \cite{IKC}             & ×4       & 31.67                        & 0.8829                        & 28.31                        & 0.7643                        & 27.37                        & 0.7192                        & 25.33                        & 0.7504                        & 28.91                        & 0.8782                        \\
\hdashline
DANv1 \citep{DANv1}          &       & 31.83                        & 0.8868                        & 28.40                        & 0.7687                        & 27.82                        & 0.7457                        & 25.05                        & 0.7365                        & 29.78                        & 0.8910                        \\
DANv2 \citep{DANv2}          &       & 31.96                        & 0.8891                        & 28.48                        & 0.7715                        & 27.86                        & 0.7484                        & 25.13                        & 0.7386                        & 29.73                        & 0.8911                        \\
DCLS \citep{DCLS}          &       & 32.07                        & 0.8893                        & 28.52                        & 0.7727                        & 27.89                        & 0.7488                        & 25.34                        & 0.7465                        & 30.07                        & 0.8968                        \\
NLCUNet \citep{NLCUNet}&       & \textbf{32.29} & \textbf{0.8931} & \underline{28.66} & \underline{0.7751} & \underline{27.96} & \underline{0.7518} & \textbf{25.69} & \textbf{0.7597} & \underline{30.16} & \underline{0.9004}\\

BSRGAN \cite{bsrgan}  &       & 27.47 & 0.7913 & 25.64 & 0.6638 & 25.64 & 0.6510 & 22.35 & 0.6133 & 25.00 & 0.7785\\

Real-ESRGAN \cite{Real-ESRGAN}  &       & 26.16 & 0.7919 & 24.91 & 0.6661 & 25.11 & 0.6553 & 21.38 & 0.6002 & 24.71 & 0.7882\\

LabNet (Ours)  &       & \underline{32.27} & \underline{0.8923} & \textbf{28.70} & \textbf{0.7769} & \textbf{27.98} & \textbf{0.7536} & \underline{25.62} & \underline{0.7583} & \textbf{30.27} & \textbf{0.9011}\\ \hline

\end{tabular}%
}
\end{table*}

\begin{table*}[] \tiny
\centering
\caption{The ablation experiments of the STEA module. We compared the effects of first-order Taylor expansion approximation (FTEA) and second-order Taylor expansion approximation (STEA) on the performance, parameters, and MFLOPs of SR (with scale ×4). The test datasets were Set5, Urban100, and Manga109. The best and second-best results are \textbf{highlighted} and \underline{underlined}.}
\label{tab:STEA_abla}
\resizebox{0.7\textwidth}{!}{%
\begin{tabular}{cccccccccccccccc}
\hline
DWC(V)                         & FTEA                & STEA    & \multicolumn{2}{c}{Set5}                    & \multicolumn{2}{c}{Urban100}                & \multicolumn{2}{c}{Manga109}                 \\
                            &                             &                                                                               & PSNR                 & SSIM                                 & PSNR                 & SSIM                 & PSNR                 & SSIM                 \\ \hline
\XSolidBrush & \Checkmark   & \XSolidBrush                                                   & 31.97                & 0.8886                              & 25.19                & 0.7417               & \underline{29.94}                & 0.8946              \\
\Checkmark   & \Checkmark   & \XSolidBrush                                                  & \underline{32.04}                & \underline{0.8891}                              & \underline{25.20}                & \underline{0.7429}               & 29.93               & \underline{0.8954}               \\
\Checkmark   & \XSolidBrush & \Checkmark                                                   & \textbf{32.07}                & \textbf{0.8896}                              & \textbf{25.31}                & \textbf{0.7462}               & \textbf{29.98}                & \textbf{0.8964}                 \\
 \hline
\end{tabular}%
}
\end{table*}

\begin{table}[t] \tiny
\centering
\caption{Comparison of MFLOPs and Parameters in Different configurations.}
\label{tab:STEA_MFLOPs}
\resizebox{0.45\textwidth}{!}{%
\begin{tabular}{cccccccc}
\hline
DWC(V)                         & FTEA                 & STEA   & TTEA              & \# Params               & MFLOPs               \\ \hline
\XSolidBrush & \Checkmark   & \XSolidBrush  & \XSolidBrush                                                  & 10.6K                & 33                   \\
\Checkmark   & \Checkmark   & \XSolidBrush  & \XSolidBrush                                                   & 16.4K                & 63                   \\
\Checkmark   & \XSolidBrush & \Checkmark   & \XSolidBrush                                                     & 19.1K                & 71                   \\ 
\Checkmark   & \XSolidBrush & \XSolidBrush & \Checkmark                                                       & 21.8K                & 80                   \\ 
 \hline
\end{tabular}%
}
\end{table}

\begin{table}[t] \tiny
\centering
\caption{In the case of using DWC(V) and STEA, comparison of MFLOPs and parameters for MLFR in different configurations.}
\label{tab:MLFR_MFLOPs}
\resizebox{0.45\textwidth}{!}{%
\begin{tabular}{cccccccc}
\hline
\begin{tabular}[c]{@{}c@{}}MLFR\_v1\end{tabular} & \begin{tabular}[c]{@{}c@{}}MLFR\_v2\end{tabular} & \begin{tabular}[c]{@{}c@{}}MLFR\_v3\end{tabular} & \# Params               & MFLOPs               \\ \hline
\Checkmark                                         & \XSolidBrush                                             & \XSolidBrush                                                  & 50.5K                & 198                  \\
\XSolidBrush                                       & \Checkmark                                               & \XSolidBrush                                                  & 64.1K & 251 \\
\XSolidBrush                                       & \XSolidBrush                                             & \Checkmark                                                    & 89.9K & 358 \\ \hline
\end{tabular}%
}
\end{table}

\begin{figure*}[]
\centering
\includegraphics[scale=0.48]{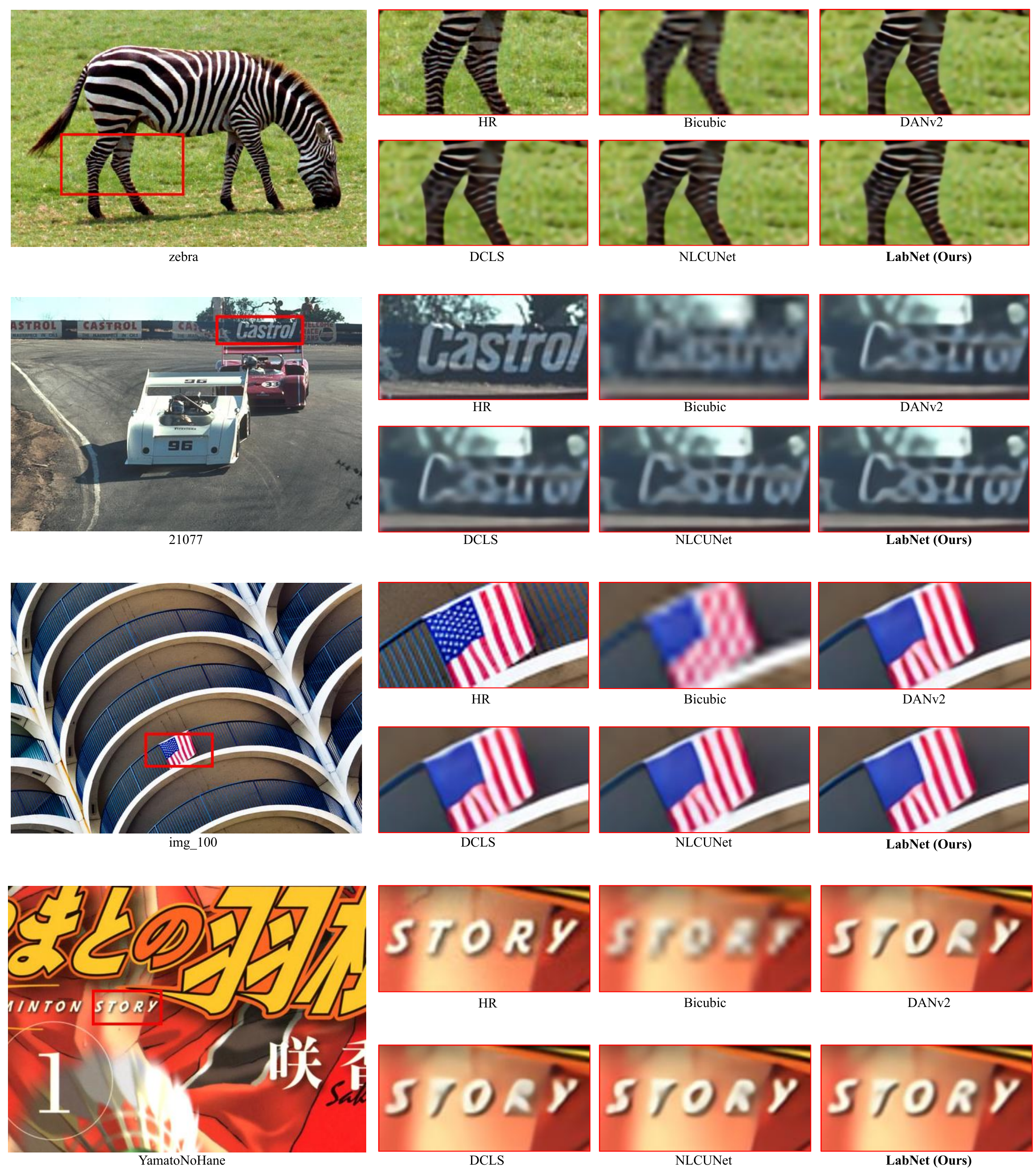}
\vspace{-0.3cm}
\caption{Visual results are from Set14, BSD100, Urban100 and Manga109 at scale x4. Here, the width of the blur kernel is 3.2.}
\label{fig:visual_result}
\end{figure*}

\begin{figure*}\tiny
\centering
\includegraphics[scale=0.4]{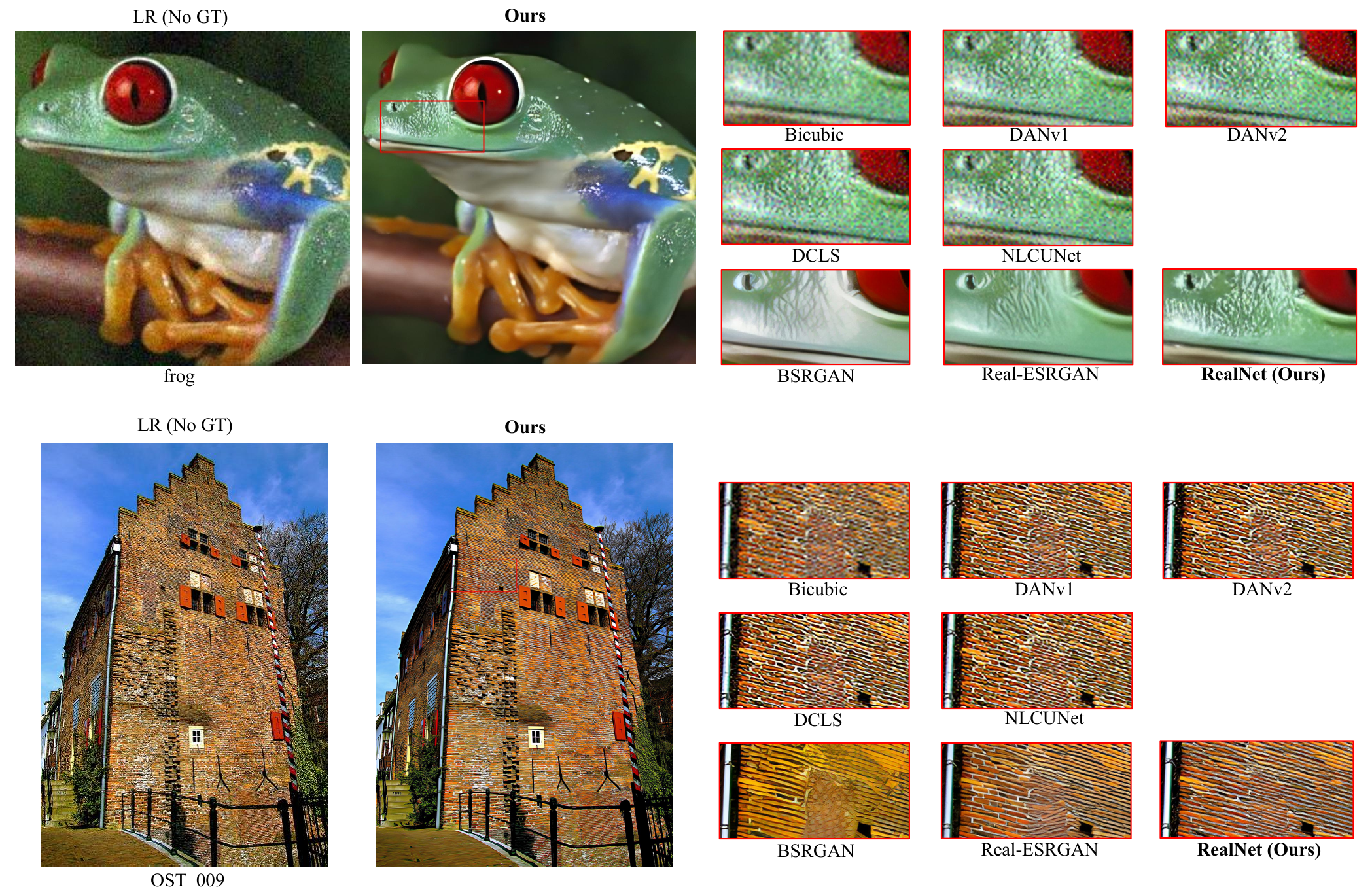}
\caption{\small The visualization results consist of various methods, including the non-deep learning approach Bicubic, as well as the DANv1, DANv2, DCLS, and NLCUNet, which utilize only Gaussian blurring and downsampling during training. Additionally, the visualization includes the Real-ESRGAN, BSRGAN, and our RealNet, which involve training on datasets with complex degradations.}
\label{fig:real-compare}
\end{figure*}

\begin{table*}[] \tiny
\centering
\caption{The ablation experiments were conducted on the MLFR module, using DWC(V) and STEA, where MLFR\_v1 uses depth-wise convolution with a kernel size of 9 and a dilation rate of 6, MLFR\_v2 uses the same depth-wise convolution as MLFR\_v1, but also includes an additional depth-wise convolution with a kernel size of 9 and a dilation rate of 4, and MLFR\_v3 uses the same configurations as MLFR\_v2, but also includes an additional depth-wise convolution with a kernel size of 9 and a dilation rate of 2. The best and second-best results are \textbf{highlighted} and \underline{underlined}, respectively.}
\label{tab:MLFR_abla}
\resizebox{0.77\textwidth}{!}{%
\begin{tabular}{cccccccccccccccc}
\hline
 \begin{tabular}[c]{@{}c@{}}MLFR\_v1\end{tabular} & \begin{tabular}[c]{@{}c@{}}MLFR\_v2\end{tabular} & \begin{tabular}[c]{@{}c@{}}MLFR\_v3\end{tabular} & \multicolumn{2}{c}{Set5}                                                        & \multicolumn{2}{c}{Urban100}                & \multicolumn{2}{c}{Manga109}                 \\
                                                     &                                                                   &                                                                                                                          & PSNR                 & SSIM                 & PSNR                 & SSIM                                & PSNR                 & SSIM                                 \\ \hline
\Checkmark                                         & \XSolidBrush                                             & \XSolidBrush                                                   & 32.09                & 0.8907                                           & \underline{25.53}                & 0.7539               & \textbf{30.31}                & \underline{0.8998}  \\
\XSolidBrush                                       & \Checkmark                                               & \XSolidBrush                                                   & \underline{32.23} & \underline{0.8920}  & 25.50 & \underline{0.7527} & 30.25 & \underline{0.8998}  \\ 
 \XSolidBrush                                       & \XSolidBrush                                               & \Checkmark                                                   & \textbf{32.27} & \textbf{0.8923}  & \textbf{25.62} & \textbf{0.7583} & \underline{30.27} & \textbf{0.9011}  \\ \hline
\end{tabular}%
}
\end{table*}

\begin{table*}[]
\centering
\caption{Ablation experiments on the combined and individual effects of STEA and MLFR. The best and the second best results are \textbf{highlighted} and \underline{underlined}.}
\label{tab:STEA_MLFR}
\resizebox{0.76\textwidth}{!}{%
\begin{tabular}{cccccccccccc}
\hline
STEA                & MLFR & \multicolumn{2}{c}{Set5}                    & \multicolumn{2}{c}{Set14}                   & \multicolumn{2}{c}{BSD100}                  & \multicolumn{2}{c}{Urban100}                & \multicolumn{2}{c}{Manga109}                \\                                                                        &                                                                               & PSNR                 & SSIM                 & PSNR                 & SSIM                 & PSNR                 & SSIM                 & PSNR                 & SSIM                 & PSNR                 & SSIM                 \\ \hline
\Checkmark   & \XSolidBrush                                                   & 32.07                & 0.8896               & 28.51                & 0.7713               & 27.89                & 0.7481               & 25.31                & 0.7462               & 29.98                & 0.8964                   \\ 
\XSolidBrush & \Checkmark                                                   & \underline{32.10} & \underline{0.8898} & \underline{28.58} & \underline{0.7734} & \underline{27.93} & \underline{0.7489} & \underline{25.43} & \underline{0.7485} & \underline{30.14} & \underline{0.8972} \\
\Checkmark  & \Checkmark                                                   & \textbf{32.27} & \textbf{0.8923} & \textbf{28.70} & \textbf{0.7769} & \textbf{27.98} & \textbf{0.7536} & \textbf{25.62} & \textbf{0.7583} & \textbf{30.27} & \textbf{0.9011} \\ \hline
\end{tabular}%
}
\end{table*}

\begin{table}[t]
\centering
\caption{Comparison of inference time, number of parameters, and GFLOPs for RealNet in different configurations.}
\label{tab:RealNet_compare}
\resizebox{0.46\textwidth}{!}{%
\begin{tabular}{cccc}
\hline
Scale: $\times4$ & Infer\_time on GPU & \# Params & GFLOPs \\ \hline
Without NLA                                 & 0.40 s             & 4 M    & 17     \\
With NLA                                    & 4.21 s             & 5.3 M  & 194    \\
With NLSA                                   & 5.15 s             & 5.4 M  & 69     \\
With STEA                                   & 0.48 s             & 5.8 M  & 25     \\
With STEA + MLFR                        & 0.53 s                 & 11.5 M     & 47      \\ \hline 
\end{tabular}%
}
\end{table}

\subsection{\textbf{Implementation Details of LabNet}}
Our LabNet is built on the foundation of the LSTEA module, forming a U-Net. Each LSTEA module consists of multiple LSTEA blocks, as shown in Fig.~\ref{fig:LabNet}, the number of LSTEA blocks in each LSTEA module from left to right is: \{$6$, $8$, $10$, $6$, $8$, $8$\}. During the training process, we set the size of the input image to 64 × 64, and we use the Mean Absolute Error (MAE) as the loss function. The total iterations are 1.2$\times$10$^6$. The initial learning rate is set to 4$\times$10$^{-4}$ and is halved every 3$\times$10$^5$ iterations. We use Adam \cite{adam} as the optimizer with $\beta_1$ = $0.9$ and $\beta_2$ = $0.99$. All models are trained on Tesla T4 GPUs.\\
$\textbf{Data Preprocessing of LabNet.}$ For the training dataset, a blur operation with a blur kernel size of 21 × 21 is applied. For different scaling factors (×2, ×3, and ×4), the variance of the blur kernel is uniformly sampled during the training process from the ranges [0.2, 2.0], [0.2, 3.0], and [0.2, 4.0], respectively. For the test dataset, we uniformly select 8 blur kernels from the ranges [0.80, 1.60], [1.35, 2.40], and [1.8, 3.2] for different scaling factors (×2, ×3, and ×4), respectively.

\subsection{\textbf{Implementation Details of RealNet}}
Our RealNet is constructed by two branches: denoising and deblurring. As illustrated in Fig.~\ref{fig:RealNet}, both the denoising branch and the deblurring branch consist of 4 modules, each composed of 5 LSTEA modules. The entire network is trained for 30 epochs. The initial learning rate is set to $1\times 10^{-3}$ and is halved every 15 epochs. We use Adam as the optimizer with $\beta_1$ = $0.9$ and $\beta_2$ = $0.99$. All models are trained on Tesla T4 GPUs.\\
$\textbf{Data Preprocessing of RealNet.}$ As shown in Fig.~\ref{fig:train_test}, we first independently and in parallel apply second-order noise and second-order blur processing to the HR image, resulting in two corresponding LR images. Then, these two LR images are passed through the denoising and deblurring branches of RealNet, respectively. Finally, a fusion operation is performed to obtain the restored SR image.


\subsection{\textbf{Comparisons with State-of-the-art}}
\noindent $\textbf{Quantitative Evaluations in Lab Scenarios.}$ From Table~\ref{tab:sota_compare}, it can be observed that our LabNet significantly outperforms other state-of-the-art deep SR models on nearly all benchmarks and scaling factors. For instance, compared to NLCUNet \cite{NLCUNet} with a scale factor of ×2, our LabNet achieves performance improvements of 0.03 dB, 0.05 dB, 0.11 dB, and 0.02 dB on the Set14, BSD100, Urban100, and Manga109 datasets, respectively. Similarly, compared to NLCUNet with a scale factor of ×4, our LabNet achieves performance improvements of 0.04 dB, 0.02 dB, and 0.11 dB on the Set14, BSD100, and Manga109 datasets, respectively. Our LabNet consistently outperforms DCLS \cite{DCLS} across all scale factors. Additionally, as shown in Table~\ref{tab:sota_GFLOPs}, when the input image size is (1,3,64,64) and the scaling factor is ×4, our LabNet achieves an inference time of 1.18 seconds on a GPU, which is 1.05 seconds faster than NLCUNet but slightly slower than DCLS. Furthermore, our LabNet has significantly fewer GFLOPs compared to NLCUNet and is comparable to DCLS. In summary, our LabNet not only achieves comparable performance to NLCUNet but also exhibits higher efficiency. This can be attributed to the utilization of LSTEA in our LabNet, which combines STEA and MLFR to capture multi-scale features with a large receptive field while maintaining linear complexity.

\noindent $\textbf{Qualitative Evaluations in Lab Scenarios.}$ As shown in Fig.~\ref{fig:visual_result}, by comparing the reconstruction result of the $'$zebra$'$ in Fig.~\ref{fig:visual_result}, we can observe that the black and white stripes on the legs of the zebra generated by our LabNet are close to the GT image. Furthermore, by comparing the reconstruction result of $'$YamatoNoHane$'$, we can observe that our LabNet performs better in recovering the textual content in the image compared to other methods.

\noindent $\textbf{Quantitative Evaluation in Real-World Scenarios.}$ As commonly observed, GAN-based SR methods often generate lower values in terms of PSNR and SSIM metrics while producing visually captivating results. As shown in Table~\ref{tab:sota_compare}, Real-ESRGAN and BSRGAN perform significantly lower than non-GAN-based SR methods on the five benchmark test datasets. However, Fig.~\ref{fig:real-compare} reveals the impressive visual outcomes achieved by Real-ESRGAN and BSRGAN. Additionally, due to the absence of corresponding GT images for the degraded images in the RealWorld38 dataset, we only conducted visual comparisons between our RealNet and other blind SR methods.

\noindent $\textbf{Qualitative Evaluation in Real-World Scenarios.}$ The experimental results are presented in Fig.~\ref{fig:real-compare}. We evaluated the generalization capability of our model using the RealWorld38 dataset, which consists of real-world images. Our RealNet outperforms previous methods for reconstructing LR images in real-world scenarios. Among all the methods, the non-deep learning bicubic interpolation method performs the worst, followed by DANv2, DCLS, and NLCUNet. These methods only employ Gaussian blur and downsampling during training, resulting in visually unsatisfactory outcomes. BSRGAN \cite{bsrgan} and Real-ESRGAN achieve favorable visual results due to their training on datasets with complex degradations. However, upon comparing the $'$OST\_009$'$ image in Fig.~\ref{fig:real-compare}, we observe that the result obtained by BSRGAN exhibits a significant color deviation from the LR input, while the result produced by Real-ESRGAN evokes a strong sense of discordance. In contrast, our RealNet reconstruction result closely resembles the color of the input LR image and demonstrates improved overall coherence.


\subsection{\textbf{Ablation Study and Analysis of LabNet}}
\subsubsection{\textbf{Impact of STEA}}
Table~\ref{tab:STEA_abla} showcases the effect of employing first-order Taylor expansion approximation (FTEA) and second-order Taylor expansion approximation (STEA) on the performance of blind SR tasks.
By comparing the PSNR and SSIM values of the first and second rows in Table~\ref{tab:STEA_abla}, we observed an improvement in performance across all test datasets when DWC($\cdot$) was applied to the Value. Furthermore, comparing the PSNR and SSIM values of the second and third rows in Table~\ref{tab:STEA_abla}, we found that utilizing STEA for the exponential function outperformed FTEA across all test datasets. Particularly on the Urban100 dataset, the PSNR was higher by 0.11 dB when using a STEA compared to a FTEA. Furthermore, we found that when the order of Taylor expansion is greater than 2, the performance does not show a significant improvement. \\
$\textbf{Efficiency Analysis.}$ As shown in Table~\ref{tab:STEA_MFLOPs}, when applying DWC($\cdot$) to Value, there is a slight increase in the number of parameters and a slight rise in MFLOPs. By comparing rows 2 to 4 in Table \ref{tab:STEA_MFLOPs}, as the order of Taylor expansion increases, the third-order Taylor expansion approximation (TTEA) parameter count and MFLOPs are greater than STEA, which are greater than FTEA. Ultimately, we used STEA to build our LabNet and RealNet.

\subsubsection{\textbf{Impact of MLFR}}
Table~\ref{tab:MLFR_abla} highlights the influence of incorporating multi-scale dilated convolutions in MLFR on blind SR performance. By comparing the PSNR and SSIM values of the first to third rows in Table~\ref{tab:MLFR_abla}, we found that in MLFR, when utilizing dilated convolutions with dilations of 2, 4, and 6 simultaneously, the performance was the best across all test datasets. Particularly on the Urban100 dataset, the PSNR was 0.12 dB higher when using dilated convolutions with dilations of 2, 4, and 6 compared to dilated convolutions with dilations of 4 and 6. \\
$\textbf{Efficiency Analysis.}$ As shown in Table~\ref{tab:MLFR_MFLOPs}, in MLFR, the maximum number of parameters and MFLOPs are achieved when utilizing dilated convolutions with Dilated = 6, 4, and 2 simultaneously.

\subsubsection{\textbf{Impact of STEA and MLFR}}
As shown in Table~\ref{tab:STEA_MLFR}, LabNet achieves the best performance on all five benchmarks when both STEA and MLFR are used simultaneously.

\subsection{\textbf{Ablation Study and Analysis of RealNet}}
The experimental results from Table~\ref{tab:RealNet_compare} highlight the following insights: Our RealNet without NLA has the fastest inference speed, processing a 64 × 64 input image in just 0.40 seconds. In contrast, Our RealNet with NLA and NLSA have much higher computational costs, with inference times of 4.21 seconds and 5.15 seconds, respectively. Our RealNet with the STEA, achieves high efficiency by processing an input image in only 0.48 seconds. Our RealNet with the STEA and the MLFR strikes a favorable balance between efficiency and visual result, processing an input image in 0.53 seconds.

\subsection{\textbf{Details of adjustable parameters in RealNet}}
\noindent \textbf{Adjusting a single parameter:} Users can adjust the values of the denoising and deblurring parameters in RealNet to achieve the desired effects. In the visualization shown in Fig.~\ref{fig:single_params}, Increasing $\alpha_1$ slightly enhances image details, but excessive increase may lead to distorted edges. Gradually increasing $\alpha_3$ improves the distinction between objects and the background, but setting it too high can negatively impact the image. $\alpha_2$ progressively enhances image visualization as it increases, but setting it to extremely large values can damage the image. Careful adjustment of $\alpha_4$ is necessary to prevent color distortion, as it is related to the deep layer. It is advisable to avoid setting $\beta_1$ to $\beta_4$ to excessively large values to prevent color distortion in the resulting image. \\
\textbf{Joint modification of multiple parameters:} Users can still fine-tune the denoising and deblurring ratios (i.e., $\alpha_i$ and $\beta_i$) according to their preferences, even after the training process. In the visualization shown in Fig.~\ref{fig:multi_params}, the second and third rows display the restored results using different ratios for denoising and deblurring. This demonstrates that RealNet is capable of restoring rich texture information from LR images, catering to different user preferences.

\section{Conclusion}
In this paper, we presented a second-order Taylor expansion approximation (STEA) to handle the high computational complexity of self-similarity techniques. Then, we designed a multi-scale large-field reception (MLFR) to compensate for the super-resolution (SR) performance loss caused by STEA. To confirm the superiority of our designed STEA and MLFR, we respectively constructed the LabNet for testing in laboratory scenarios and the RealNet for testing in real-world scenarios. Extensive experimental results demonstrated the effectiveness of our method in blind SR tasks. The ablation study further validates their contribution to the overall framework.
\bibliographystyle{IEEEbib}
\bibliography{spl}





\end{document}